# A New Constructive Method to Optimize Neural Network Architecture and Generalization

Hou Muzhou[a,*], Moon Ho Lee[b],

[a] School of Mathematics and Statistics, Central South University, Changsha 410083, China (e-mail: houmuzhou@sina.com).
[b] College of Engineering, Chonbuk National University, Korea

*Abstract*—**In this paper, after analyzing the reasons of poor generalization and overfitting in neural networks, we consider some noise data as a singular value of a continuous function - jump discontinuity point. The continuous part can be approximated with the simplest neural networks, which have good generalization performance and optimal network architecture, by traditional algorithms such as constructive algorithm for feed-forward neural networks with incremental training, BP algorithm, ELM algorithm, various constructive algorithm, RBF approximation and SVM. At the same time, we will construct RBF neural networks to fit the singular value with every $\varepsilon$ error in $L^2(\mathbb{R}^d)$, and we prove that a function with $m$ jumping discontinuity points can be approximated by the simplest neural networks with a decay RBF neural networks in $L^2(\mathbb{R})$ by each $\varepsilon$ error, and a function with $m$ jumping discontinuity point $y = f(x), \quad x \in E \subset \mathbb{R}^d$ can be constructively approximated by a decay RBF neural networks in $L^2(\mathbb{R}^d)$ by each $\varepsilon > 0$ error and the constructive part have no generalization influence to the whole machine learning system which will optimize neural network architecture and generalization performance, reduce the overfitting phenomenon by avoid fitting the noisy data.**

## I. INTRODUCTION

Neural Networks have attracted increasing attention from researchers in many fields, including information processing, computer science, economics, medicine and mathematics, and have been used to solve a wide range of problems such as data mining, function approximation, pattern recognition, expert system and data prediction etc. The widespread popularity of neural networks in many fields is mainly due to their ability to approximate complex multivariate nonlinear functions directly from the input samples. Neural networks can provide models for a large class of natural and artificial phenomena that are difficult to handle using classical parametric techniques.

One of the most important problems that neural network designers face today is choosing an appropriate network size for a given application. However, the process of selecting adequate neural network architecture for a given problem is still a controversial issue.

\* Corresponding author. Tel: +86 13787088322.
E-mail address: houmuzhou@sina.com (Hou Muzhou),

And it is empirically known that the problem is particularly serious when the size of the network is large. When a network is trained with noisy data, it may have a very small training error that is caused by fitting the noise rather than the true function underlying the data. In such situations, the generalization error tends to be larger than its optimal level because the trained network may deviate from the true function[1]. We also call this phenomenon overfitting [2-5].

The overfitting problem is a critical issue that usually leads to poor generalization [6-8]. One of the main reasons of overfitting is the excessive noise data or singular value in the practical problems [9, 10]. So the traditional methods of processing noise data is to remove noise data before approximation through various algorithms such as wavelet transform [11-14], principal component analysis [15, 16] and various filtering algorithms [17-20]. But sometimes, the "noisy" data we think of ways to remove is often some singular value of a process which contains important information [21].

In this paper, we will consider some noise data as a singular value of a continuous function - jump discontinuity point. The continuous part can be approximated using less size neural network, which have optimal architecture and good generalization performance, by traditional algorithms such as constructive algorithm for feedforward neural networks with incremental training [22, 23], BP algorithm [24, 25], ELM algorithm [26, 27], various constructive algorithm[28-30], RBF approximation [31-33] and SVM [34]. At the same time, we will construct a RBF neural network to fit the singular value with every $\varepsilon$ error in $L^2(\mathbb{R}^d)$, and then the whole network will have optimal architecture and generalization.

This paper is organized as follows; Section 2 investigates the phenomenon of neural network overfitting caused by noisy data. Section 3 gives some previous works on approximation of functions by neural networks. Section 4 investigates constructive multidimensional approximation of a function with one jump discontinuity point. Theorems 6-9 are proved. Section 5 investigates constructive multidimensional approximation of a function with finite number of jump discontinuity point, Theorems 10-13 are proved. Section 6 provides some conclusions.

## II. THE PHENOMENON OF NEURAL NETWORKS OVER-FITTING CAUSED BY NOISY DATA

Overfitting is a well-known generalization problem for neural network due to the finite training set, which greatly reduce its generalization ability in practical applications [35]. But one of the important factors to cause overfitting is noisy data[36]. In the real world, some underlying function relationship between the input and desired output are simple, but the sample data are always corrupted by noise to some degree. Consequently learning with noisy data would need too many hidden neurons and then results in poor generalization ability. We will describe this phenomenon through a simple experiment.

Example: the following sample dataset $A$ comes from the function $y = \cos x$ which contained 60 points (Fig.1). And sample dataset $B$ comes from function

$$y = f_1(x) = \begin{cases} \cos x & x \neq \pi \\ -2 & x = \pi \end{cases},$$

which has a noise data in function $y = \cos x$ and also contained 60 points (Fig.2).

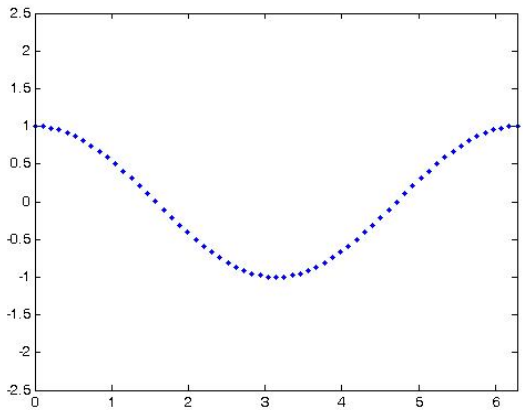

Fig.1 sample dataset $A$ comes from function $y = \cos x$

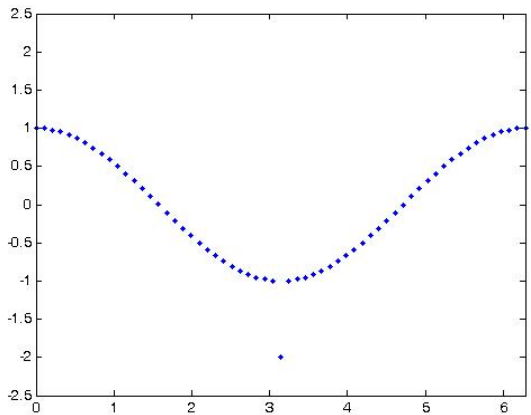

Fig.2 sample dataset $B$ comes from function $y = f_1(x)$

In the following experiments, we will approximate the dataset $A$ and $B$ using traditional single hidden layer feedforward neural networks trained by BP algorithm. Although there are many variants of BP algorithm, a faster BP algorithm called Levenberg-Marquardt is used in our experiments. All the experiments are carried out in MATLAB 7.10 (R2010a) environment running in a Intel(R) Core(TM) i3-2120 3.30GHz CPU. We will use 54 points of the dataset to train the neural networks and the other 6 points to test the networks.

In the experiments, we will compare the following performance index: training time; the ratio of the training time with the training time of dataset $A$ (RTT)

$$RTT = \frac{training\ time\ of\ dataset\ B}{training\ time\ of\ dataset\ A};$$

training and testing root mean square error (RMSE)

$$RMSE = \sqrt{\frac{1}{n+1}\sum_{i=0}^{n}(f(x_i) - \tilde{y}_i)^2}$$

where $\tilde{y}_i$ is the output value of the neural networks in the simulation point; the ratio of training RMSE (RTRR)

$$RTRR = \frac{trainging\ time\ of\ dataset\ B}{training\ time\ of\ dataset\ A}$$

the ratio of testing RMSE (RTER)

$$RTER = \frac{testing\ time\ of\ dataset\ B}{testing\ time\ of\ dataset\ A}$$

and the number of the hidden neurons.

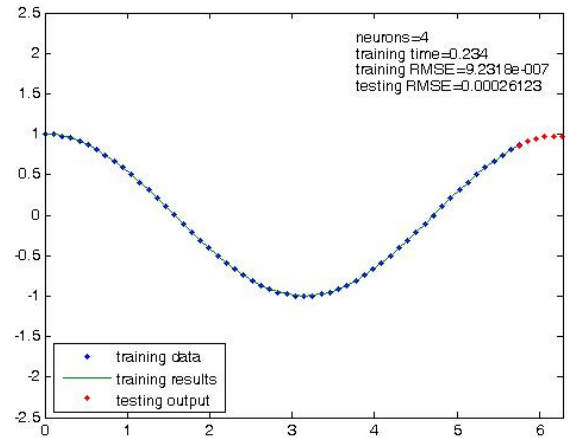

Fig.3 training and testing to dataset $A$ with 4 neurons

In Fig.3, the neural network only need 4 hidden neurons to fit the function $y = \cos x$ very well with 0.234 CPU training time, $9.2318 \times 10^{-7}$ training RMSE and 0.00026123 testing RMSE.

TABLE I
PERFORMANCE COMPARISON FOR DATASET $A$ WITH $B$

| Performance<br>Dataset | Training | | | | Testing | | Neurons |
|---|---|---|---|---|---|---|---|
| | Time | RTT | RMSE | RTRR | RMSE | RTER | |
| $A$ | 0.234 | 1 | $9.2318 \times 10^{-7}$ | 1 | 0.00026123 | 1 | 4 |
| $B$ | 5.8968 | 25.2 | 0.012537 | 13580 | 0.0057322 | 21.943 | 4 |
| $B$ | 2.4024 | 10.267 | $9.9419 \times 10^{-7}$ | 1.0769 | 0.0008318 | 3.2085 | 10 |

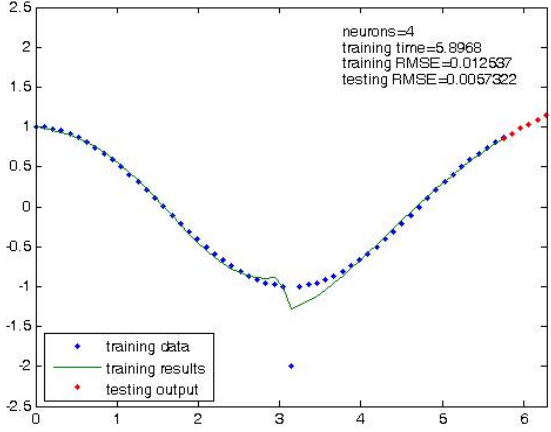

Fig.4 training and testing to dataset $B$ with 4 neurons

In Fig.4, because there is one noise point in dataset $B$, 4 hidden neurons in the neural networks is not enough to fit the dataset $B$, which it spent 5.8926 CPU time to train the networks with 0.012537 training RMSE that is 13580 times of that in Fig.3, and 0.0057322 testing RMSE which is 21.943 times of that in Fig.3.

So we increased hidden neurons to 10 to fit it well, in Fig 5, it spent 2.4024 CPU time to train the networks with $9.9419 \times 10^{-7}$ training RSME which is almost equals that in Fig.1, but the testing RMSE is 0.00083815 that is larger 3.2085 times than that in Fig.1. So with the increase of the complexity of the networks and the precision of training RMSE, the generalization of the neural networks was reduced instead because of the noise data.

The results above three experiments are concluded into the table 1.

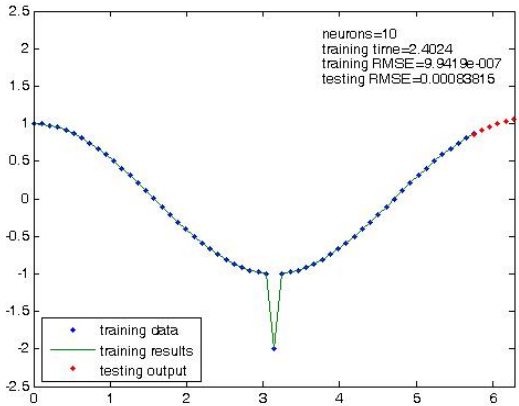

Fig.5 training and testing to dataset $B$ with 10 neurons

In this paper, the dataset with noisy data is divided into two parts, one part contains simple function relationship, and the other part consists of jumping discontinuity points. The first part can be approximated with the optimal neural network architecture, which has less number of hidden neurons and good generalization performance, by traditional algorithm such as BP, ELM and SVM. At the same time, we will construct a RBF neural network to approximate the singular value with every $\varepsilon$ error in $L^2(\mathbb{R}^d)$ which has no influence to the generalization of the first part and the whole neural networks.

### III. PREVIOUS WORKS ON APPROXIMATION OF FUNCTIONS BY NEURAL NETWORKS

There are many good results on approximation of continuous functions relationship without noisy data by neural networks that have best performing architecture and well generalization by properly using traditional algorithm such BP, ELM, SVM and some constructive approaches.

Let $\phi : \mathbb{R} \to \mathbb{R}$. Define

$$\sum\nolimits^d (\phi) \equiv span\{\phi(\mathrm{w}\cdot\mathrm{x}+b) : b \in \mathbb{R},\ \mathrm{w},\mathrm{x} \in \mathbb{R}^d\}, \quad (1)$$

Then $N \in \sum^d(\phi)$ if and only if

$$N(\mathrm{x}) = \sum_{j=0}^{n} c_j \phi(\mathrm{w}_j \cdot \mathrm{x} + b_j), \quad (2)$$

where $c_j,\ j = 0,\cdots,n$ are real numbers and $n$ is a positive integer.

We say that $\phi$ is a sigmoid function, if it verifies $\lim_{t \to -\infty}\phi(t) = 0$ and $\lim_{t \to \infty}\phi(t) = 1$, then (2) is called single-layer feed-forward neural network. If $\phi(\mathrm{x}) = \varphi(\|\mathrm{x}-\mathrm{x}_0\|)$ $\mathrm{x} \in \mathbb{R}^d$, we call $\phi(\mathrm{x})$ RBF function, then (2) is called RBF neural networks.

In the case of continuous functions we have the following density results

**Theorem 1**([37]) Let $\phi \in C(\mathbb{R})$. Then $\sum^d(\phi)$ is dense in $C(\mathbb{R}^d)$ in the topology of uniform convergence on compact if and only if $\phi$ is not a polynomial.

**Corollary 1**([38]) $y = f(\mathrm{x}) \in C(\mathbb{R}^d)$ can be approximated by a simplest neural networks (such as with Minimum number of hidden neurons).

In the case of not necessarily continuous functions we also have some density results.

**Theorem 2**([39]) Let $\phi$ be bounded, measurable and sigmoidal. Then $\sum^d(\phi)$ is dense in $L^1([0,1]^d)$.

The following theorem is a generalization of the above results.

**Theorem 3**([40]) Let $\phi$ be a Lebesgue measurable function, not a.e. equal to a polynomial, satisfying $\int_a^b |\phi(x)|^p dx < \infty$ for all $a, b \in \mathbb{R}$. Let $K$ be a compact set in $\mathbb{R}^d$. Then for any function $f \in L^p(K)$ ($p \geq 1$) and every $\varepsilon > 0$, there is a network $N \in \sum^d(\phi)$ such that

$$\|N - f\|_{K,p} < \varepsilon,$$

where $\|g\|_{K,p} \equiv (\int_K |g(x)|^p dx)^{\frac{1}{p}}$.

For RBF neural networks we have the following results.

**Theorem 4**([31]) Let $\phi$ be a RBF function, Then for any function $f \in L^p(\mathbb{R})$ ($p \geq 1$) and every $\varepsilon > 0$, there is a network $N \in \sum^d(\phi)$ such that

$$\|N - f\|_p < \varepsilon,$$

where $\|g\|_p \equiv (\int_\mathbb{R} |g(x)|^p dx)^{\frac{1}{p}}$.

**Theorem 5**([28]) 1. Let $x \in [a,b]^k \subset \mathbb{R}^k$, and $f(x)$ be a multivariate continuous function, $x_i \in [a,b]^k$ $i = 0,1,\cdots,n$ be an uniform grid partition to $[a,b]^k$, where $[a,b]$ is divided to $s$ equal partition, and arrange by breadth-first such that $x_0, x_1, \cdots, x_n$ ($n = s^k$). Then

$$\|x_i - x_{i-1}\| = \frac{b-a}{n^{\frac{1}{2k}}} = \frac{b-a}{s^{\frac{1}{2}}}.$$

2. $A$ depends on $n$, that is $A = A(n)$.

3. The real numbers $f_i$ are the images of $x_i$ under a multivariate continuous function $f(x)$, that is $f_i = f(x_i)$, $i = 0,1,2,\cdots,n$.

For each $\varepsilon > 0$, we can construct a decay RBF neural network $W_a(x, A(n))$, and there exists a function $A(n)$ and a natural number $N$ such that, when $n > N$, we have

$$|f(x) - W_a(x, n, A(n))| < \varepsilon,$$

for all $x \in [a,b]^k$.

The above theorems on approximation of continuous function can be carried out by many traditional machine learning systems with good generalization such as in Fig.3.

## IV. CONSTRUCTIVE MULTIDIMENSIONAL APPROXIMATION OF A FUNCTION WITH ONE JUMP DISCONTINUITY POINT

In this section, we introduce a new constructive approach to reduce the overfitting phenomenon of machine learning system, especially neural networks, which can reduce the hidden neurons to optimal neural network architecture. The sample dataset with one noisy data is divided into two parts, the first part is considered to come from a simple continuous function. The other part is consisted of noisy data and is considered as one jumping discontinuous point of the continuous function. Then we can use many traditional methods to fit the continuous function with optimal architecture and good generalization, for the noisy part, we can construct a decay RBF neural network to fit it without influence the generalization to the first part and the whole machine learning system.

**Definition 1**. Consider $\psi(x)$ be a continuous real function, also the condition is given as $\lim_{x \to \infty} \psi(x) = 0 = o(e^{-x^2})$, and $\psi(0) \neq 0$. We call $\psi(x)$ is a decay RBF, and the decay RBF neural networks (DRNNs) can be written as

$$NW(x) = \sum_{j=0}^n c_j \psi(\lambda_j \|x - t_j\|), \quad x, t_j \in R^k, \quad (3)$$

where $\lambda_j, t_j$ are inner weights, $c_j$ outer weights respectively, and for $L^2(\mathbb{R})$, $\|x\| = (\int_R |x(t)|^2 dt)^{\frac{1}{2}}$

As we known that Gaussian function $\psi(x) = e^{-x^2}$ and wavelet functions, e.g. Mexican Hat wavelet $\psi(x) = (2/\sqrt{3})\pi^{-1/4}(1-x^2)e^{-x^2/2}$ and Morlet wavelet $\psi(x) = (2/\sqrt{3})e^{-x^2/2}\cos 5x$ belong to decay RBF. The definition and nature of wavelet function are introduced in [41, 42].

Replace $\varphi(x) = \psi(x)/\psi(0)$ to (3), we can get $\varphi(0) = 1$, $\lim_{x \to \infty} \varphi(x) = 0$ then we can rewriting (3) as follow

$$NW(x) = \sum_{j=0}^n c_j \psi(\lambda_j \|x - t_j\|) = \sum_{j=0}^n c_j \psi(0) \frac{\psi(\lambda_j \|x - t_j\|)}{\psi(0)} = \sum_{j=0}^n k_j \varphi(\lambda_j \|x - t_j\|)$$

**Lemma 1**. Consider $\varphi(x)$ is a decay RBF function, there exist real numbers $k_1$, $k_2$ and positive real number $A$, such that when $|x| > A$, we have $k_1 e^{-x^2} < \varphi(x) < k_2 e^{-x^2}$, and $\varphi(x)$ is bounded in $\mathbb{R}$

**Proof**, as $\lim_{x \to \infty} \varphi(x) = 0 = o(e^{-x^2})$, then $\lim_{x \to \infty} \frac{\varphi(x)}{e^{-x^2}} = k$, and for each $\varepsilon > 0$ there exists a positive real number $A$, such that when $x > A$, we have $\left|\frac{\varphi(x)}{e^{-x^2}} - k\right| < \varepsilon$, that is, $-\varepsilon < \frac{\varphi(x)}{e^{-x^2}} - k < \varepsilon$, then $k - \varepsilon < \frac{\varphi(x)}{e^{-x^2}} < k + \varepsilon$ and

$$k_1 e^{-x^2} = (k-\varepsilon)e^{-x^2} < \varphi(x) < (k+\varepsilon)e^{-x^2} = k_2 e^{-x^2}. \square$$

Now, consider $y = f(x)$ $x \in \mathbb{R}^d$ is continuous except $x = x_0$, that is meaning, $\lim_{x \to x_0} f(x) \neq f(x_0)$, We can decompose $f(x)$ as two parts $f_c(x)$ and $f_d(x)$, it`s shown as $f(x) = f_c(x)$, $+ f_d(x)$ where

$$f_c(x) = \begin{cases} f(x) & x \neq x_0 \\ \lim_{x \to x_0} f(x) & x = x_0 \end{cases} \text{ is continuous and}$$

$$f_d(x) = \begin{cases} 0 & x \neq x_0 \\ f(x_0) - \lim_{x \to x_0} f(x) = h_0 & x = x_0 \end{cases}$$

Example 1: for one-dimensional function:
$$y = f(x) = \begin{cases} \cos x & x \neq \pi \\ -2 & x = \pi \end{cases}$$

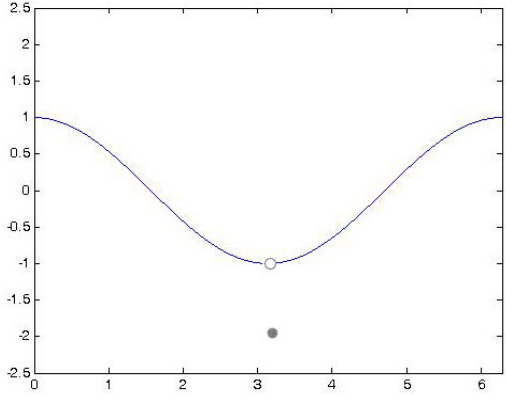

Fig.6 the figure of function $y = f(x)$

$f(x)$ can be decomposed as to $f_c(x)$ and $f_d(x)$, that is meaning $f(x) = f_c(x) + f_d(x)$, where

$$f_c(x) = \cos x \quad \text{and} \quad f_d(x) = \begin{cases} 0 & x \neq \pi \\ -1 & x = \pi \end{cases}$$

Example 2: 2-D function $z = f(x, y)$

$$z = \begin{cases} \dfrac{\sin\sqrt{x^2+y^2}}{\sqrt{x^2+y^2}} & x^2 + y^2 \neq 0 \\ 2 & x^2 + y^2 = 0 \end{cases}$$

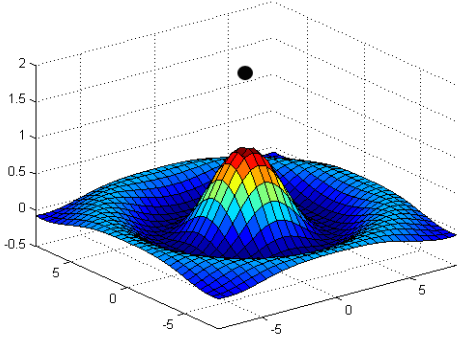

Fig.7 the figure of function $z = f(x, y)$

We can decompose $f(x, y)$ to $f_c(x, y)$ and $f_d(x, y)$, so $f(x, y)$ can be denoted as $f(x, y) = f_c(x, y) + f_d(x, y)$, where

$$f_c(x, y) = \begin{cases} f(x, y) & x^2 + y^2 \neq 0 \\ \lim_{x^2+y^2 \to 0} f(x, y) = 1 & x^2 + y^2 = 0 \end{cases} \text{ is continuous and}$$

$$f_d(x, y) = \begin{cases} 0 & x^2 + y^2 \neq 0 \\ f(0,0) - \lim_{x^2+y^2 \to 0} f(x, y) = 1 = h_0 & x^2 + y^2 = 0 \end{cases}$$

**Theorem 6** For each $\varepsilon > 0$, there exists a constructive RBF neural networks $NW_d(x, A)$ and a positive real number $A'$, such that when $A > A'$ we have
$$\|f_d(x) - NW_d(x, A)\| < \varepsilon$$

**Proof,** consider $NW_d(x, A) = h_0 \varphi(A\|x - x_0\|)$ is a RBF neural networks with one neuron, so

$$NW_d(x_0, A) = h_0 \varphi(A\|x - x_0\|) = h_0$$
$$\|f_d(x) - NW_d(x, A)\| = \|NW_d(x, A)\|$$
$$= (\int_R |h_0 \varphi(A\|x - x_0\|)|^2 dx)^{\frac{1}{2}}$$
$$= |h_0|(\int_R |\varphi(A\|x - x_0\|)|^2 dx)^{\frac{1}{2}}$$
$$= |h_0|[(\int_{\|x-x_0\|<\delta} |\varphi(A\|x - x_0\|)|^2 dx)^{\frac{1}{2}}$$
$$+ (\int_{\|x-x_0\|\geq\delta} |\varphi(A\|x - x_0\|)|^2 dx)^{\frac{1}{2}}]$$

By Lemma 1, $|\varphi(x)| < M > 0$, then

$$|h_0|[(\int_{\|x-x_0\|<\delta} |\varphi(A\|x - x_0\|)|^2 dx)^{\frac{1}{2}} < |h_0| M \delta < \frac{\varepsilon}{2}, \text{ only if } \delta < \frac{\varepsilon}{2|h_0|M},$$

and when $A\|x - x_0\| > A_1$ that is, $A > \dfrac{A_1}{\|x - x_0\|} > \dfrac{A_1}{\delta}$, we can get

$$|h_0|(\int_{\|x-x_0\|\geq\delta} |\varphi(A\|x - x_0\|)|^2 dx)^{\frac{1}{2}}$$
$$< |h_0| \int_R |k_2| e^{-A^2 x^2} dx < |h_0||k_2| \int_R e^{-A^2 x^2} dx,$$
$$= \frac{|h_0||k_2|}{A} \int_R e^{-x^2} dx = \frac{|h_0||k_2|}{A} \sqrt{\pi} < \varepsilon$$

Then, when $A > \dfrac{|h_0||k_2|}{\varepsilon} \sqrt{\pi}$, we can get

$$\|f_d(x) - NW_d(x, A)\| < \frac{\varepsilon}{2} + \frac{\varepsilon}{2} = \varepsilon. \square$$

**Remark 1**, the decay neural network $NW_d(x, A)$ can be constructed to fit the noisy data without influence to the generalization of whole machine learning.

**Theorem 7** A function with one jumping discontinuity point can be repaired to a continuous function in $L^2(\mathbb{R})$ by decay RBF neural networks with each $\varepsilon$ error.

**Poof,** It is obvious by Theorem 6. $\square$

For Example1, function
$$y = \begin{cases} \cos x & x \neq \pi \\ -2 & x = \pi \end{cases}$$

can be repaired by $\tilde{y} = \cos x + e^{-A^2(x-\pi)^2}$ with Gaussian RBF function $e^{-x^2}$ for each $\varepsilon$ when $A > A_1$, having $\|y - \tilde{y}\| < \varepsilon$

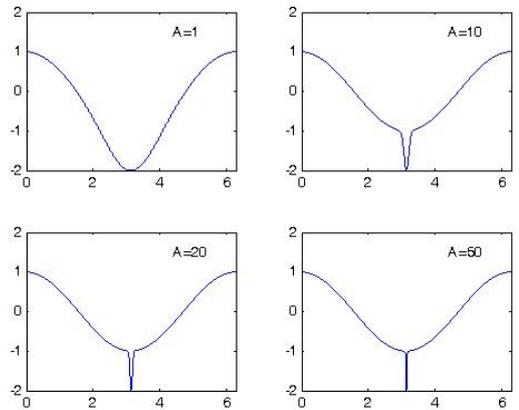

Fig.8 The repaired approximation of function $y = f(x)$ with decay RBF neural networks

Example2, The function
$$z = \begin{cases} \dfrac{\sin\sqrt{x^2+y^2}}{\sqrt{x^2+y^2}} & x^2+y^2 \neq 0 \\ 2 & x^2+y^2 = 0 \end{cases}$$

can be repaired by $\tilde{z} = f_c(x,y) + e^{-A^2(x^2+y^2)}$ with Gaussian RBF function $e^{-(x^2+y^2)}$ for each $\varepsilon$ when $A > A_1$, having $\|z - \tilde{z}\| < \varepsilon$

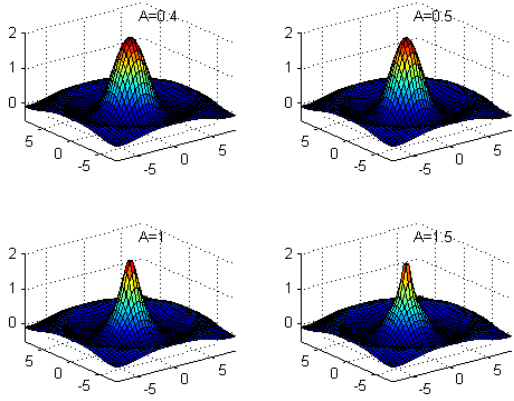

Fig.9 The repaired approximation of the function $z = f(x,y)$ with decay RBF neural networks

**Theorem 8** A function with one jumping discontinuity point can be approximated by a simplest neural networks with a decay RBF neural networks in $L^2(\mathbb{R})$ by each $\varepsilon$ error.

**Poof.** It is obvious by theorem 4 and Corollary 1

**Theorem 9** A function with one jumping discontinuity point $y = f(x)$, $x \in E \subset \mathbb{R}^d$ can be constructively fitted by a decay RBF neural networks in $L^2(\mathbb{R}^d)$ by each $\varepsilon > 0$ error.

**Poof,** As $f(x) = f_c(x) + f_d(x)$, $f_c(x) \in E \subset C(\mathbb{R}^d)$, from theorem 4, for each $\varepsilon > 0$, we can construct a decay RBF neural network $NW_c(x,A)$, and there exists a natural number $A_1$ such that, when $A > A_1$, we have
$$|f_c(x) - NW_c(x,A)| < \frac{\varepsilon}{2},$$
for all $x \in E \subset \mathbb{R}^d$.

And from Theorem 6 For each $\varepsilon > 0$, there exist a constructive RBF neural networks $NW_d(x,A)$ and a positive real number $A_2$, such that when $A > A_2$ we have
$$\|f_d(x) - NW_d(x,A)\| < \frac{\varepsilon}{2}.$$

Then when $A > A_0 = \max\{A_1, A_2\}$ we have
$$\begin{aligned}
&|f(x) - (NW_c(x,A) + NW_d(x,A))| \\
&\leq |f_c(x) - NW_c(x,A)| + |f_d(x) - NW_d(x,A)| \\
&< \frac{\varepsilon}{2} + \frac{\varepsilon}{2} = \varepsilon
\end{aligned} \quad (4)$$

And $NW(x,A) = NW_c(x,A) + NW_d(x,A)$ is also a constructive RBF neural networks.□

Based on the above theorems, a sample dataset with one noisy point can be divided into two parts; the first part can be considered from a simple continuous function which can be fit by traditional machine learning system such as Fig.1, the noisy part can be fit by a decay neural network with constructive approach. Then the whole machine learning system has optimal architecture and well generalization of part one, which can reduced the overfitting phenomenon of machine learning system.

## V. CONSTRUCTIVE MULTIDIMENSIONAL APPROXIMATION OF FUNCTION WITH FINITE NUMBER OF JUMP DISCONTINUITY POINT

The content of this section is the extension of Section IV to the situation of $m$ noisy points which has the same conclusion of optimizing neural network architecture and generalization.

Let $y = f(x)$ $x \in \mathbb{R}^d$ has $m$ jumping discontinuity points $x_j$ $j = 1,2,\cdots,m$, that is, $\lim_{x \to x_j} f(x) \neq f(x_j)$ $j = 1,2,\cdots,m$. We can decompose $f(x)$ to $f_c(x)$ and $f_d(x)$, that is, $f(x) = f_c(x) + f_d(x)$, where

$$f_c(x) = \begin{cases} f(x) & x \neq x_j \quad j=1,2,\cdots,m \\ \lim_{x \to x_j} f(x) & x = x_j \quad j=1,2,\cdots,m \end{cases}, \quad (5)$$

is continuous and

$$f_d(x) = \begin{cases} 0 & x \neq x_j \quad j=1,2,\cdots,m \\ f(x_j) - \lim_{x \to x_j} f(x) = h_j & x = x_j \quad j=1,2,\cdots,m \end{cases} \quad (6)$$

**Theorem 10** If $y = f(x)$ $x \in \mathbb{R}^d$ has $m$ jumping discontinuity points $x_j$ $j = 1,2,\cdots,m$, that is,
$$\lim_{x \to x_j} f(x) \neq f(x_j) \quad j = 1,2,\cdots,m$$
and $f(x) = f_c(x) + f_d(x)$ such as (5) and (6), for each $\varepsilon > 0$, there exist a constructive RBF neural networks $NW_d(x,A)$ and a positive real number $A'$, such that when $A > A'$ we have
$$\|f_d(x) - NW_d(x,A)\| < \varepsilon$$

**Proof,** Firstly let $f_d(x) = \sum_{j=1}^{m} f_{dj}(x)$, where
$$f_{dj}(x) = \begin{cases} 0 & x \neq x_j \\ f(x_j) - \lim_{x \to x_j} f(x) = h_j & x = x_j \end{cases}.$$

Then for each jumping discontinuity point $x_j$, by theorem 6, for each $\varepsilon > 0$, there exist a constructive RBF neural networks $NW_{dj}(x,A) = h_j \varphi(A\|x - x_j\|)$ with one neuron and a positive real number $A'_j$, such that when $A > A'_j$ we have
$$\|f_{dj}(x) - NW_{dj}(x,A)\| < \frac{\varepsilon}{m}.$$

Then, we construct
$$NW_d(x,A) = \sum_{j=1}^{m} NW_{dj}(x,A) = \sum_{j=1}^{m} h_j \varphi(A\|x - x_j\|),$$
when $A > A' = \max\{A'_1, A'_2, \cdots, A'_m\}$, we have

$$\|f_d(\mathrm{x}) - NW_d(\mathrm{x}, A)\| \leq \sum_{j=1}^{m} \|f_{dj}(\mathrm{x}) - NW_{dj}(\mathrm{x}, A)\|$$

$$< m\frac{\varepsilon}{m} = \varepsilon \qquad \square$$

**Theorem 11** A function with $m$ jumping discontinuity point can be repaired to a continuous function in $L^2(\mathbb{R})$ by decay RBF neural networks with each $\varepsilon$ error.

**Poof,** It is obvious by Theorem 10.

**Theorem 12** A function with $m$ jumping discontinuity point can be approximated by a simplest neural networks and a decay RBF neural networks in $L^2(\mathbb{R})$ by each $\varepsilon$ error.

**Poof,** It can be proved similarly to Theorem 8.

**Theorem 13** A function with $m$ jumping discontinuity point $y = f(\mathrm{x})$, $\mathrm{x} \in E \subset \mathbb{R}^d$ can be constructively approximated by a decay RBF neural networks in $L^2(\mathbb{R}^d)$ by each $\varepsilon > 0$ error.

**Poof,** It can be proved similarly to theorem 9.

## VI. Conclusions

In this paper, in order to optimize neural network architecture and generalization, after analyzing the reasons of overfitting and poor generalization of the neural networks, we presented a class of constructive decay RBF neural networks to repair the singular value of a continuous function with finite number of jumping discontinuity points. We proved that a function with $m$ jumping discontinuity points can be approximated by a simplest neural network and a decay RBF neural network in $L^2(\mathbb{R})$ by each $\varepsilon$ error, and a function with $m$ jumping discontinuity point $y = f(\mathrm{x})$, $\mathrm{x} \in E \subset \mathbb{R}^d$ can be constructively approximated by a decay RBF neural network in $L^2(\mathbb{R}^d)$ by each $\varepsilon > 0$ error. Then the whole networks will have less hidden neurons and well generalization in the same of the first part.

## Acknowledgment

This work was supported by the World Class University (WCU, R32-2011-000-20014-0) and Fundamental Research (FR, 20100020942) sponsored by National Research Foundation (NRF), Republic of Korea.

## References

[1] Hagiwara, K. and K. Fukumizu, *Relation between weight size and degree of over-fitting in neural network regression.* Neural Networks, 2008. **21**(1): p. 48-58.
[2] Yu, Z., et al. *The Design of RBF Neural Networks for Solving Overfitting Problem*: IEEE.
[3] Lawrence, S. and C.L. Giles. *Overfitting and neural networks: conjugate gradient and backpropagation*. 2000: IEEE.
[4] Tetko, I.V., D.J. Livingstone, and A.I. Luik, *Neural network studies. 1. Comparison of overfitting and overtraining.* Journal of chemical information and computer sciences, 1995. **35**(5): p. 826-833.
[5] Schittenkopf, C., G. Deco, and W. Brauer, *Two strategies to avoid overfitting in feedforward networks.* Neural Networks, 1997. **10**(3): p. 505-516.
[6] Perrone, M.P. and L.N. Cooper, *When networks disagree: Ensemble methods for hybrid neural networks.* 1992, DTIC Document.
[7] Tay, F. and L. Cao, *Application of support vector machines in financial time series forecasting.* OMEGA-OXFORD-PERGAMON PRESS-, 2001. **29**: p. 309-317.
[8] Doan, C. and S. Liong. *Generalization for multilayer neural network: Bayesian regularization or early stopping.* 2004.
[9] Liu, J., O. Demirci, and V.D. Calhoun, *A parallel independent component analysis approach to investigate genomic influence on brain function.* Signal Processing Letters, IEEE, 2008. **15**: p. 413-416.
[10] Soltani, S., *On the use of the wavelet decomposition for time series prediction.* Neurocomputing, 2002. **48**(1): p. 267-277.
[11] Prasad, G.K. and J. Sahambi. *Classification of ECG arrhythmias using multi-resolution analysis and neural networks.* 2003: IEEE.
[12] Shyu, L.Y., Y.H. Wu, and W. Hu, *Using wavelet transform and fuzzy neural network for VPC detection from the Holter ECG.* Biomedical Engineering, IEEE Transactions on, 2004. **51**(7): p. 1269-1273.
[13] He, Y., Y. Tan, and Y. Sun. *Wavelet neural network approach for fault diagnosis of analogue circuits.* 2004: IET.
[14] Chen, B., et al., *Application of wavelets and neural networks to diagnostic system development, 1, feature extraction.* Computers & chemical engineering, 1999. **23**(7): p. 899-906.
[15] Karhunen, J., et al., *A class of neural networks for independent component analysis.* Neural Networks, IEEE Transactions on, 1997. **8**(3): p. 486-504.
[16] Hyvärinen, A. and E. Oja, *Independent component analysis: algorithms and applications.* Neural Networks, 2000. **13**(4-5): p. 411-430.
[17] Shah, S., F. Palmieri, and M. Datum, *Optimal filtering algorithms for fast learning in feedforward neural networks.* Neural Networks, 1992. **5**(5): p. 779-787.
[18] Connor, J.T., R.D. Martin, and L. Atlas, *Recurrent neural networks and robust time series prediction.* Neural Networks, IEEE Transactions on, 1994. **5**(2): p. 240-254.
[19] Feraund, R., et al., *A fast and accurate face detector based on neural networks.* Pattern Analysis and Machine Intelligence, IEEE Transactions on, 2001. **23**(1): p. 42-53.
[20] Haykin, S.S., *Neural networks and learning machines.* Vol. 3. 2009: Prentice Hall.
[21] Fan, L., et al., *Singular points detection based on zero-pole model in fingerprint images.* Pattern Analysis and Machine Intelligence, IEEE Transactions on, 2008. **30**(6): p. 929-940.
[22] Islam, M.M., X. Yao, and K. Murase, *A constructive algorithm for training cooperative neural network ensembles.* Neural Networks, IEEE Transactions on, 2003. **14**(4): p. 820-834.
[23] Liu, D., T.S. Chang, and Y. Zhang, *A constructive algorithm for feedforward neural networks with incremental training.* IEEE TRANSACTIONS ON CIRCUITS AND SYSTEMS PART 1 FUNDAMENTAL THEORY AND APPLICATIONS, 2002. **49**(12): p. 1876-1879.
[24] Haykin, S. and N. Network, *A comprehensive foundation.* Neural Networks, 2004. **2**.
[25] Simon, H., *Neural networks: a comprehensive foundation.* 1999: Prentice Hall.
[26] Huang, G.B., Q.Y. Zhu, and C.K. Siew. *Extreme learning machine: a new learning scheme of feedforward neural networks.* 2004: Ieee.
[27] Huang, G.B. and L. Chen, *Convex incremental extreme learning machine.* Neurocomputing, 2007. **70**(16-18): p. 3056-3062.
[28] Hou, M. and X. Han, *Constructive approximation to multivariate function by decay RBF neural network.* IEEE Transactions on Neural Networks, 2010. **21**(9): p. 1517-1523.
[29] Llanas, B. and F. Sainz, *Constructive approximate interpolation by neural networks.* Journal of computational and applied mathematics, 2006. **188**(2): p. 283-308.
[30] Islam, M.M., et al., *A new constructive algorithm for architectural and functional adaptation of artificial neural networks.* Systems, Man, and Cybernetics, Part B: Cybernetics, IEEE Transactions on, 2009. **39**(6): p. 1590-1605.
[31] Park, J. and I.W. Sandberg, *Universal approximation using radial-basis-function networks.* Neural computation, 1991. **3**(2): p. 246-257.
[32] Er, M.J., et al., *Face recognition with radial basis function (RBF) neural networks.* Neural Networks, IEEE Transactions on, 2002. **13**(3): p. 697-710.
[33] Mai-Duy, N. and T. Tran-Cong, *Approximation of function and its derivatives using radial basis function networks.* Applied Mathematical Modelling, 2003. **27**(3): p. 197-220.

[34] Chang, C.C. and C.J. Lin, *LIBSVM: a library for support vector machines.* ACM Transactions on Intelligent Systems and Technology (TIST), 2011. **2**(3): p. 27.
[35] Kai, W., et al. *An Expanded Training Set Based Validation Method to Avoid Overfitting for Neural Network Classifier.* in *Natural Computation, 2008. ICNC '08. Fourth International Conference on.* 2008.
[36] Liu, Z.P. and J.P. Castagna. *Avoiding overfitting caused by noise using a uniform training mode.* in *Neural Networks, 1999. IJCNN '99. International Joint Conference on.* 1999.
[37] Pinkus, A., *Approximation theory of the MLP model in neural networks.* Acta Numerica, 1999. **8**(1): p. 143-195.
[38] Mulero-Martínez, J.I., *Best approximation of Gaussian neural networks with nodes uniformly spaced.* Neural Networks, IEEE Transactions on, 2008. **19**(2): p. 284-298.
[39] Cybenko, G., *Approximation by superpositions of a sigmoidal function.* Mathematics of Control, Signals, and Systems (MCSS), 1989. **2**(4): p. 303-314.
[40] Burton, R.M. and H.G. Dehling, *Universal approximation in p-mean by neural networks1.* Neural Networks, 1998. **11**(4): p. 661-667.
[41] Chui, C.K., *An introduction to wavelets.* Vol. 1. 1992: Academic Pr.
[42] Delyon, B., A. Juditsky, and A. Benveniste, *Accuracy analysis for wavelet approximations.* Neural Networks, IEEE Transactions on, 1995. **6**(2): p. 332-348.